\documentclass[11pt]{article}
\usepackage{eacl2017}
\usepackage{times}
\usepackage{url}
\usepackage{color}
\usepackage{latexsym}
\usepackage{graphicx}
\usepackage{gnuplottex}
\usepackage{amsmath}
\graphicspath{ {figures/} }

\usepackage{pdfsync}
\usepackage{graphicx}
\usepackage{booktabs}
\usepackage{amssymb}
\usepackage{multicol}
\usepackage{watermark}
\usepackage{bm}
\usepackage{upgreek}
\usepackage{multirow}

\eaclfinalcopy 

\newcommand{\toggle}[1]{}
\renewcommand{\toggle}[1]{#1} 

\newif\ifcomment\commenttrue
\ifcomment
\newcommand{\macomment}[1]{\marginpar{\begin{center}\textcolor{red}{#1}\end{center}}}

\else
\newcommand{\macomment}[1]{} 
\fi

\def\textit#1{{\it #1}}
\def\textbf#1{{\bf #1}}
\def\textsl#1{{\sl #1}}
\def\texttt#1{{\tt #1}}

\newcommand{\citet}{\newcite}
\newcommand{\citep}{\cite}

\title{Vocabulary Selection Strategies for Neural Machine Translation}

\author{Gurvan L'Hostis \\
  \'Ecole polytechnique \\
  Palaiseau, France \\\And
  David Grangier \\
  Facebook AI Research \\
  Menlo Park, CA \\\And
  Michael Auli \\
  Facebook AI Research \\
  Menlo Park, CA}

\date{}

\begin{document}

\maketitle

\begin{abstract}
Classical translation models constrain the space of possible outputs 
by selecting a subset of translation rules based on the input sentence.
Recent work on improving the efficiency of neural translation models 
adopted a similar strategy by restricting the output vocabulary to a 
subset of likely candidates given the source.
In this paper we experiment with context and embedding-based selection 
methods and extend previous work by examining speed and accuracy trade-offs
in more detail.
We show that decoding time on CPUs can be reduced by up to $90$\% and 
training time by $25$\% on the WMT15 English-German and WMT16 English-Romanian 
tasks at the same or only negligible change in accuracy. 
This brings the time to decode with a state of the art neural translation
system to just over $140$ msec per sentence on a single CPU core for English-German.
\end{abstract}

\section{Introduction}


Neural Machine Translation (NMT) has made great progress in recent 
years and improved the state of the art on several benchmarks~\cite{%
jean:wmt15,sennrich:wmt16,zhou:baidu}. 
However, neural systems are typically less efficient than traditional
phrase-based translation models 
(PBMT; Koehn et al. 2003)\nocite{koehn:2003:naacl}, 
both at training and decoding time. 

The efficiency of neural models depends on the size of the target vocabulary 
and previous work has shown that vocabularies of well over 50k word types 
are necessary to achieve good accuracy~\cite{jean:nmt_large_vocab,zhou:baidu}.
Neural translation systems compute the probability of the next target 
word given both the previously generated target words as well as the 
source sentence. 
Estimating this conditional distribution is linear in the size 
of the target vocabulary which can be very
large for many language pairs~\cite{grave:softmax}.
Recent work in neural translation has adopted sampling techniques from 
language modeling which do not leverage the input sentence~\cite{mikolov:2011:asru,jean:nmt_large_vocab,chen:lm,zhou:baidu}. 

On the other hand, classical translation models generate outputs
in an efficient two-step \emph{selection} procedure: first, a subset of promising 
translation rules is chosen by matching rules to the source sentence, and by 
pruning them based on local scores such as translation probabilities. 
Second, translation hypotheses are generated that incorporate non-local
scores such as language model probabilities.

Recently, \citet{mi:manipulation} proposed a similar strategy for neural
translation: a selection method restricts the target vocabulary 
to a small subset, specific to the input sentence. 
The subset is then scored by the neural model.
Their results demonstrate that vocabulary subsets that are only 
about 1\% of the original size result in very little to no degradation
in accuracy.

This paper complements their study by experimenting with additional selection 
techniques and by analyzing speed and accuracy in more detail.
Similar to \citet{mi:manipulation}, we consider selecting target words 
based either on a dictionary built from Viterbi word alignments, or 
by matching phrases in a traditional phrase-table, or by using the $k$ 
most frequent words in the target language.
In addition, we investigate methods that do not rely on a traditional 
phrase-based translation model or alignment model to select target words.
We investigate bilingual co-occurrence counts, bilingual embeddings as 
well as a discriminative classifier to leverage context information
via features extracted from the entire source sentence 
(\textsection\ref{sec:strategies}).

Our experiments show speed-ups in CPU decoding by up to a 
factor of $10$ at very little degradation in accuracy.
Training speed on GPUs can be increased by a factor of $1.33$.
We find that word alignments as the sole selection method is sufficient 
to obtain good accuracy. This is in contrast to \citet{mi:manipulation} who used 
a combination of the $2,000$ most frequent words, word alignments as well as 
phrase-pairs.

Selection methods often fall short in retrieving all words of
the gold standard human translation. 
However, we find that with a reduced vocabulary of $\sim 600$ words 
they can recover over $99$\% of the words that are actually chosen 
by a model that decodes over the entire vocabulary.
Finally, the speed-ups obtained by vocabulary selection become even more 
significant if faster encoder models are used, since selection removes
the burden of scoring large vocabularies (\textsection\ref{sec:exp}).

\section{Vocabulary Selection Strategies}
\label{sec:strategies}

This section presents different selection strategies inspired
by phrase-based translation. 
We improve on a simple word co-occurrence method by estimating 
bilingual word embeddings with Hellinger PCA and 
then by using word alignments instead of co-occurrence counts. 
Finally, we leverage richer context in the source via
bilingual phrases from a phrase-table or by using the entire 
sentence in a Support Vector Machine classifier.
\subsection{Word Co-occurrences}

This is the simplest approach we consider. We estimate
a co-occurrence table which counts how many times each source 
word $s$ co-occurs with each target word $t$ in the training bitext.
The table allows us to estimate the joint distribution $P(s, t)$.
Next, we create a list of the $k$ target words that co-occur
most with each source word, i.e., the words $t$ which maximize 
$P(s, t)$ for a given $s$. 
Vocabulary selection then simply computes the union of the 
target word lists associated with each source word in the input.
 
We were concerned that this strategy over-selects frequent
target words which have higher co-occurrence counts than
rare words, regardless of the source word. 
Therefore, we experimented with selecting target words maximizing 
point-wise mutual information (PMI) instead, i.e., 
$$
PMI(s,t) = \frac{P(s,t)}{P(s)P(t)}
$$
However, this estimate was deemed too unreliable for low $P(t)$ 
in preliminary experiments and did not perform better than just 
$P(s,t)$.

\subsection{Bilingual Embeddings}

We build bilingual embeddings by applying Hellinger Principal Component 
Analysis (PCA) to the bilingual co-occurrence count matrix 
$M_{i,j} = P(t = i| s = j)$;
this extends the work on monolingual embeddings of~\citet{lebret:pca} 
to the bilingual case.
The resulting low rank estimate of the matrix can be more 
robust for rare counts. 
Hellinger PCA has been shown to produce embeddings which perform
similarly to word2vec but at higher speed~\cite{mikolov:word2vec,gouws:bilbowa}.

For selection, the estimated co-occurrence can be used instead of 
the raw counts as described in the above section. This strategy is 
equivalent to using the low rank representation of each 
source word (source embedding, i.e., column vectors from the PCA) 
and finding the target word with the closest low rank representation 
(target embeddings, i.e., row vectors from the PCA). 

\subsection{Word Alignments}

This strategy uses word alignments learned from a bilingual 
corpus~\cite{brown:mt}. 
Word alignment introduces latent variables to model $P(t|s)$, 
the probability of source word $t$ given target word $s$. 
Latent variables indicate the source position corresponding to each 
target position in a sentence pair~\cite{koehn:smt}.

We use FastAlign, a popular reparameterization of IBM Model 
2~\cite{dyer:fastalign}. 
For each source word $s$, we build a list of the top $k$ target
words maximizing $P(t|s)$. 
The candidate target vocabulary is the union of the lists for 
all source words.

Compared to co-occurrence counts, this strategy avoids selecting 
frequent target words when conditioning on a rare source word.
Word alignments will only link a frequent target word to a 
rare source word if no better explanation is present in 
the source sentence.

\subsection{Phrase Pairs}

This strategy relies on a phrase translation table, i.e., a table 
pairing source phrases with corresponding target phrases. 
The phrase table is constructed by reading off all bilingual
phrases that are consistent with the word alignments according to
an extraction heuristic~\cite{koehn:smt}.

For selection, we restrict the phrase table to the phrases 
present in the source sentence and consider the union of the
word types appearing in all corresponding target phrases~\cite{mi:manipulation}.
Compared to word alignments, we hope this strategy to fetch 
more relevant target words as it can rely on longer source phrases 
to leverage richer source context.

\subsection{Support Vector Machines}

Support Vector Machines (SVMs) for vocabulary selection have been previously 
proposed in \cite{bangalore:svmmt}. 
The idea is to determine a target vocabulary based on the entire source sentence
rather than individual words or phrases.
In particular, we train one SVM per target word taking as input a sparse 
vector encoding the source sentence as a bag of words. 
The SVM then predicts whether the considered target word is present or absent
from the target sentence.

This classifier-based method has several advantages compared to phrase 
alignments: the input is not restricted to a few contiguous source words and 
can leverage all words in the source sentence. 
The model can express anti-correlation with negative weights, marking 
that the presence of a source word is a negative indicator for the 
presence of a target word. 
A disadvantage of this approach is that we need to feed the source sentence
to all SVMs in order to get scores, instead of just reading from a 
pre-computed table.
However, SVMs can be evaluated efficiently since (i) the features are sparse, 
and (ii) only features corresponding to words from the source sentence 
are used at each evaluation.
Finally, this framework formulates the selection of each target word as an 
independent binary classification problem which might not favor 
competition between target words.

\subsection{Common Words}

Following \citet{mi:manipulation}, we consider adding the $k$ most frequent
target words to the above selection methods.
This set includes conjunctions, determiners, prepositions and frequent verbs.
Pruning any such word through restrictive vocabulary selection 
may adversely affect the system output and is addressed by this technique.

\section{Related Work}
\label{sec:previous}

The selection of a limited target vocabulary from the source
sentence is classical topic in machine translation. 
It is often referred to as \emph{lexical selection}. 
As mentioned above, word-based and phrase-based systems perform 
implicit lexical selection by building a word or phrase table 
from alignments to constrain the possible target words.
Other approaches to lexical selection include discriminative 
models such as SVMs and Maximum Entropy models ~\cite{bangalore:svmmt}
as well as rule-based systems~\cite{tufis:lexicon,tyers:selection}. 

In the context of neural machine translation, vocabulary size has 
always been a concern. Various strategies have been proposed to
improve training and decoding efficiency. 
Approaches inspired by importance sampling reduce the vocabulary 
for training~\cite{jean:nmt_large_vocab}, byte pair encoding 
segment words into more frequent sub-units~\cite{sennrich:bpe}, 
while~\cite{luong:char} proposes to segment words into characters. 
Related work in neural language modeling is also relevant~\cite{bengio:lm,%
mnih:hsm,chen:lm}. 
One can refer to~\cite{sennrich:tutorial} for further references. 

Closer to our work, recent work~\cite{mi:manipulation} presents
preliminary results on using lexical selection techniques in an NMT 
system. 
Compared to this work, we investigate more selection methods 
(SVM, PCA, co-occurrences) and analyze the speed/accuracy
trade-offs at various operating points.
We report efficiency gains and distinguish the impact of selection 
in training and decoding. 

\section{Experiments \& Results}
\label{sec:exp}

This section presents our experimental setup, then discusses 
the impact of vocabulary selection at decoding time and then during 
training time.

\subsection{Experimental Setup}
\label{sec:expsetup}

We use a an encoder-decoder style neural machine translation 
system based on Torch.\footnote{\parbox[t][4em][s]{0.42\textwidth}{We will 
release the code with the camera ready.}}
Our encoder is a bidirectional recurrent neural network 
(Long Short Term Memory, LSTM) and an LSTM decoder with attention.
The resulting context vector is fed to an 
LSTM decoder which generates the output~\cite{bahdanau:iclr15,luong:attention}.
We use a single layer setup both in the encoder and as well as 
the decoder, each with $512$ hidden units.
Decoding experiments are run on a CPU since this is the most 
common type of hardware for inference.
For training we use GPUs which are the most common hardware for 
neural network fitting.
Specifically, we rely on $2.5$GHz Intel Xeon 5 CPUs and 
Nvidia Tesla M40 GPUs. 
Decoding times are based on a single CPU core and training times 
are based on a single GPU card.

Word alignments are computed with 
FastAlign~\cite{dyer:fastalign} in both language directions and 
then symmetrized with 'grow-diag-final-and'.
Phrase tables are computed with Moses~\cite{koehn:moses}
and we train support vector machines with SvmSgd~\cite{bottou:svmsgd}. 
We also use the Moses preprocessing scripts to tokenize 
the training data.

We experiment on two language pairs.
The majority of experiments are on WMT-15 English to German 
data~\cite{wmt:2015}; we use newstest2013 for validation and 
newstest2010-2012 as well as newstest2014,2015 to present 
final test results. 
Training is restricted to sentences of 
no more than $50$ words which results in $3.6$m sentence pairs. 
We chose the vocabulary sizes following the same methodology. 
We use the $100$k most frequent words both for the source and target 
vocabulary.
At decoding time we use a standard beam search with a beam width 
of $5$ in all experiments.
Unknown output words are simply replaced with the source word whose
attention score is largest~\cite{luong:unk}. 

We also experiment with WMT-16 English to Romanian data using
a similar setting but allowing sentences of up to $125$
words~\cite{wmt:2016}.  
Since the training set provided by WMT is limited to $600$k sentence pairs, 
we add the synthetic training data provided by \citet{sennrich:wmt16}.
This results in a total of $2.4$m sentence pairs.
Our source vocabulary comprises the $200$k most frequent words and the 
target vocabulary contains $50$k words.

\subsection{Selection for Efficient Decoding}

\begin{figure}[htb]
\centering
\includegraphics[width=\columnwidth]{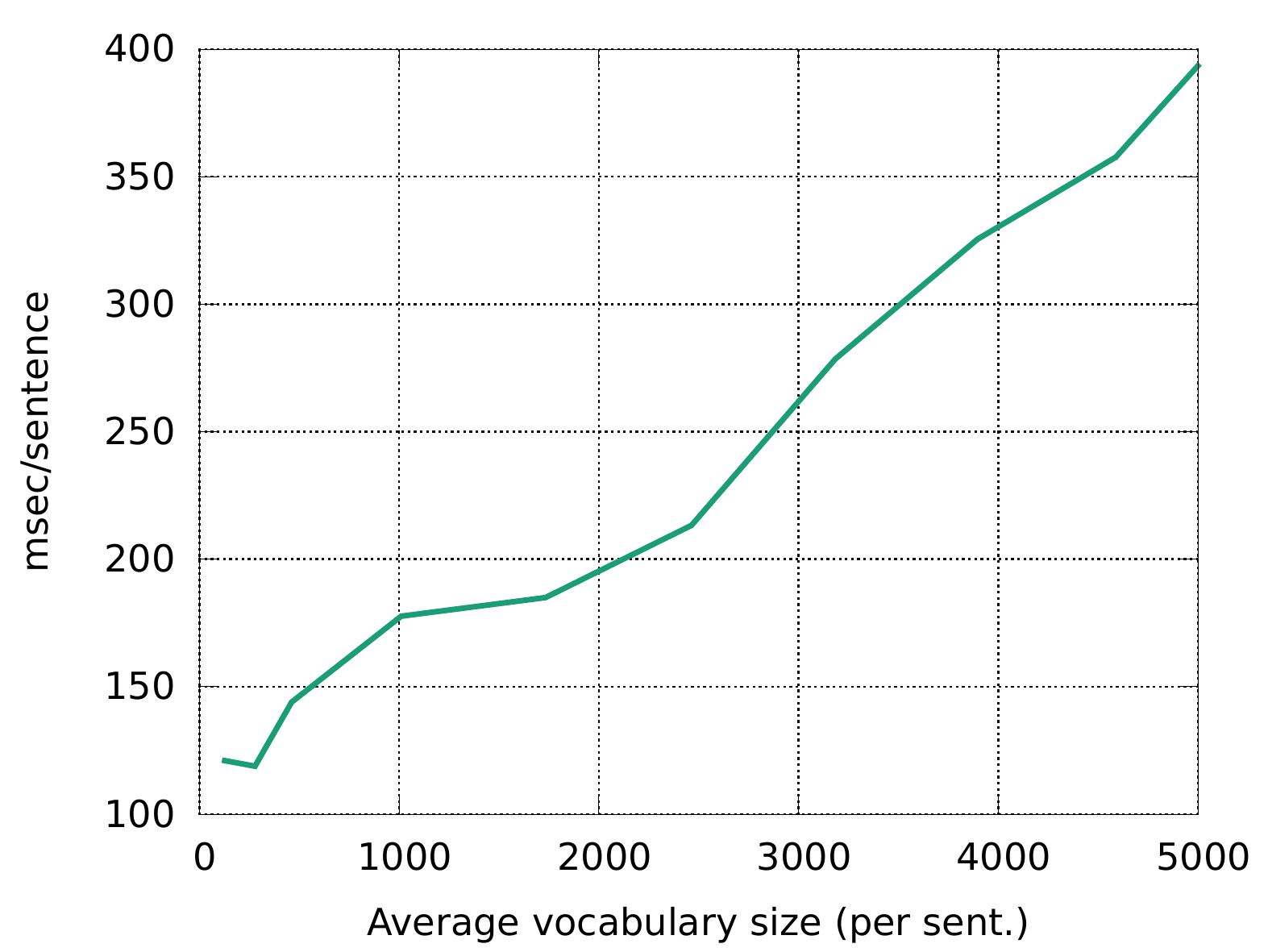}
\caption{\normalsize{Decoding time vs. vocabulary size
on newstest2013 for WMT15 English to German translation.}}
\label{fig:decoding_time_vs_vocab}
\end{figure}

Decoding efficiency of neural machine translation is still much
lower than for traditional phrase-based translation.
For NMT, the running time of beam search on a CPU is dominated by
the last linear layer that computes a score for each target word.
Vocabulary selection can therefore have a large impact on decoding speed.
Figure~\ref{fig:decoding_time_vs_vocab} shows that a reduced
vocabulary of $\sim460$ types ($144$ msec) can achieve 
a $10$X speedup over using the full 100k-vocabulary 
($\sim1,600$ msec).

\begin{figure}[htb]
\centering
\includegraphics[width=\columnwidth]{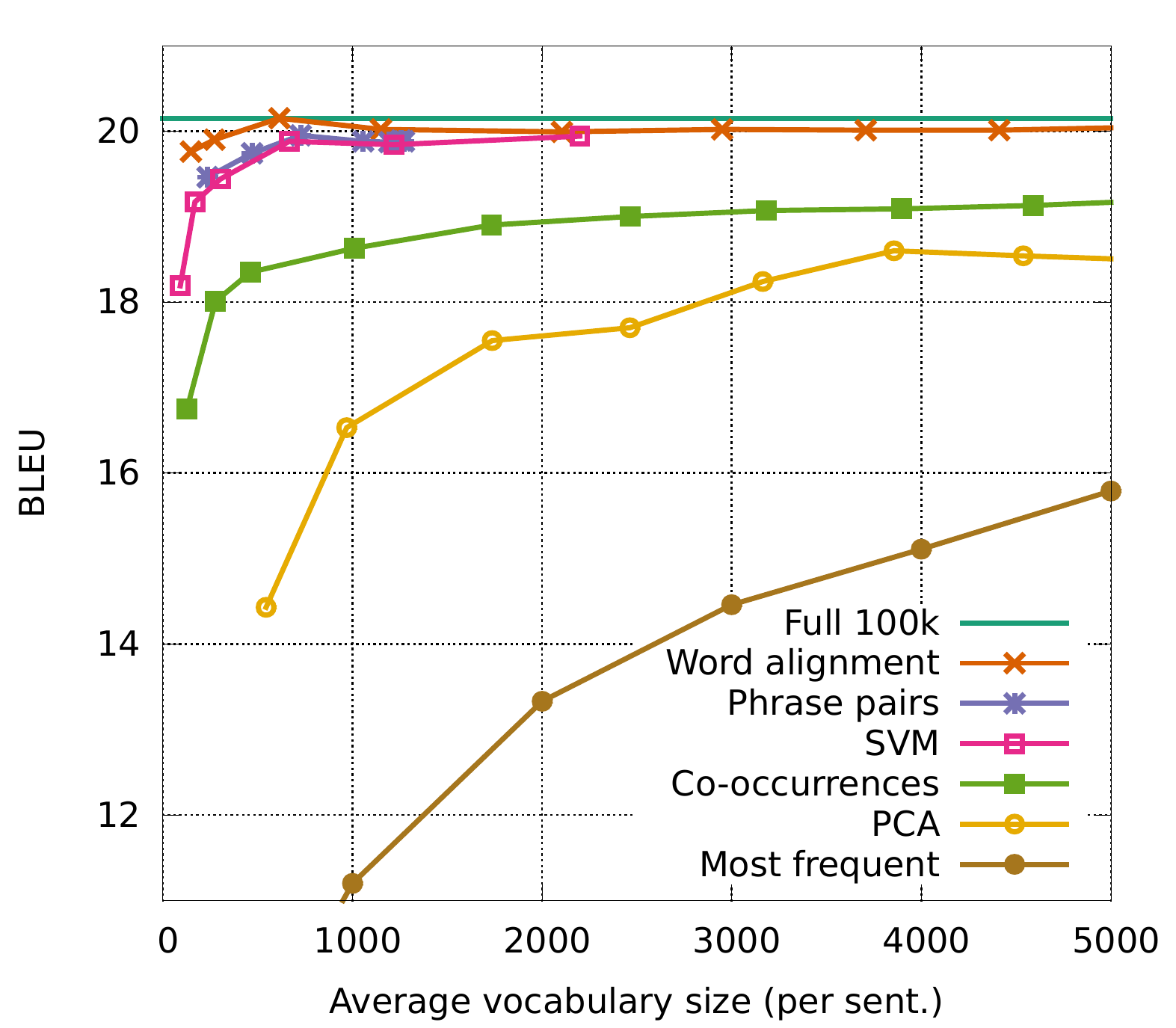}
\caption{\normalsize{BLEU vs. vocabulary size for different selection strategies.}}
\label{fig:BLEU_vs_vocab}
\end{figure}

Next we investigate the impact of reduced vocabularies on accuracy.
Figure~\ref{fig:BLEU_vs_vocab} compares BLEU for the various selection
strategies on a wide range of vocabulary sizes. 
Broadly, there are two groups of techniques: first, co-occurrence
counts and bilingual embeddings (PCA) are not able to match 
the baseline performance (Full $100$k) even with over $5$k candidate words 
per sentence. 
Second, even with fewer than $1,000$ candidates per sentence, 
word alignments, phrase pairs and SVMs nearly match the 
full vocabulary accuracy.

Although co-occurrence counts and PCA have shown useful to measure semantic
relatedness~\cite{brown:clustering,lebret:pca}, it seems that considering 
the whole source sentence as the explanation of a target word without
latent alignment variables undermines their selection ability. 
Overall, word alignments work as well or better than the other 
techniques relying on a wider input context (phrase pairs and SVMs).
Querying a word table is also more efficient than querying a phrase-table 
or evaluating SVMs. We therefore use word alignment-based selection for 
the rest of our analysis.

\begin{figure}[htb]
\centering
\includegraphics[width=\columnwidth]{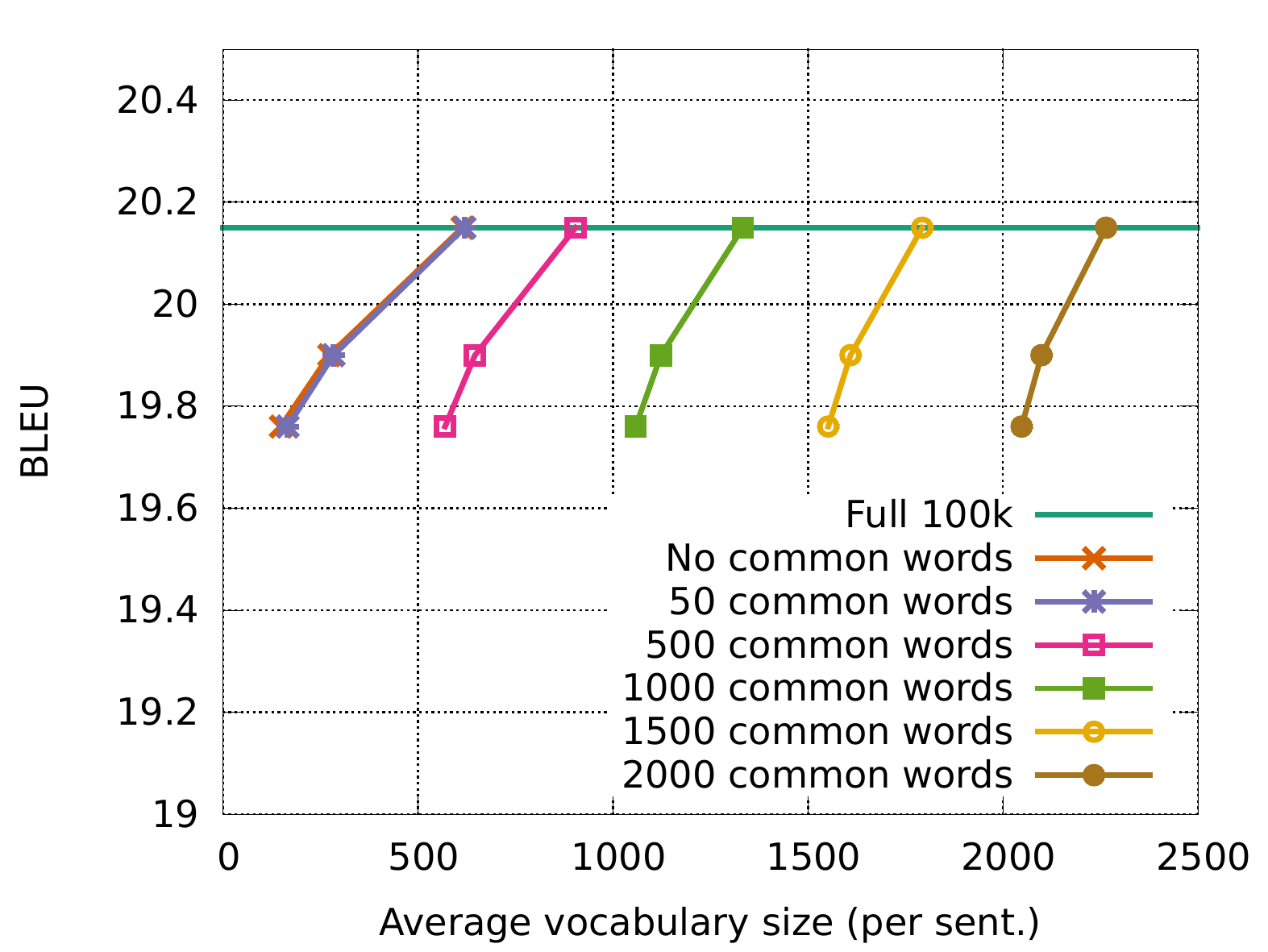}
\caption{\normalsize{Impact of adding common words to word alignment 
selection for various vocabulary sizes.}}
\label{fig:BLEU_common_vs_vocab}
\end{figure}

\citet{mi:manipulation} suggest that adding common
words to a selection technique could have a positive impact. We therefore
consider adding the most frequent $k$ words to our word alignment-based 
selection. Figure~\ref{fig:BLEU_common_vs_vocab} shows that this actually
has little impact on BLEU in our setting. In fact, the overlap of the 
results for $n=0$ and $50$ indicates that most of the top 50 words 
are already selected, even with small candidate sets.

\begin{figure*}[htb]
\centering
\includegraphics[width=\columnwidth]{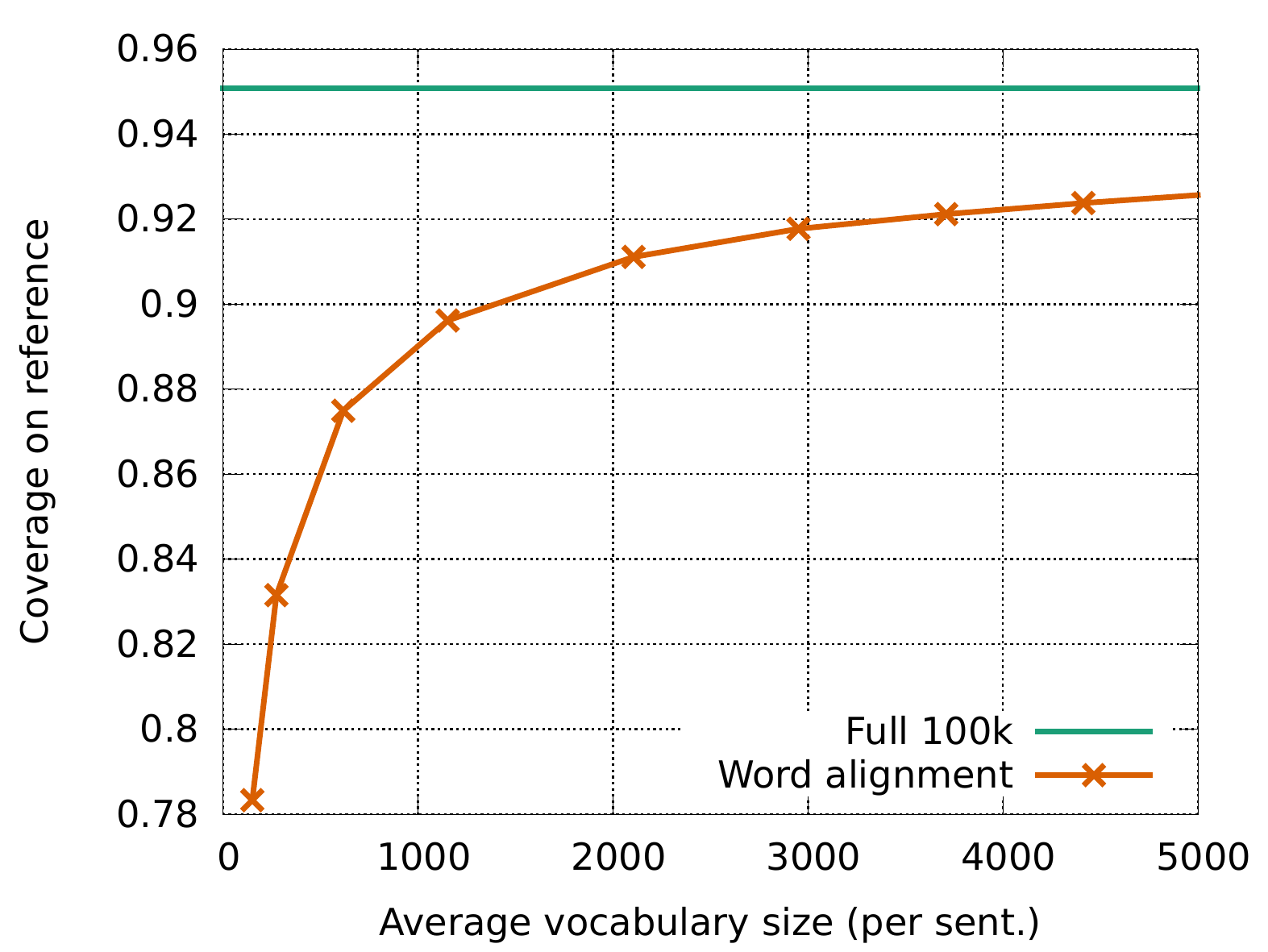}
\includegraphics[width=\columnwidth]{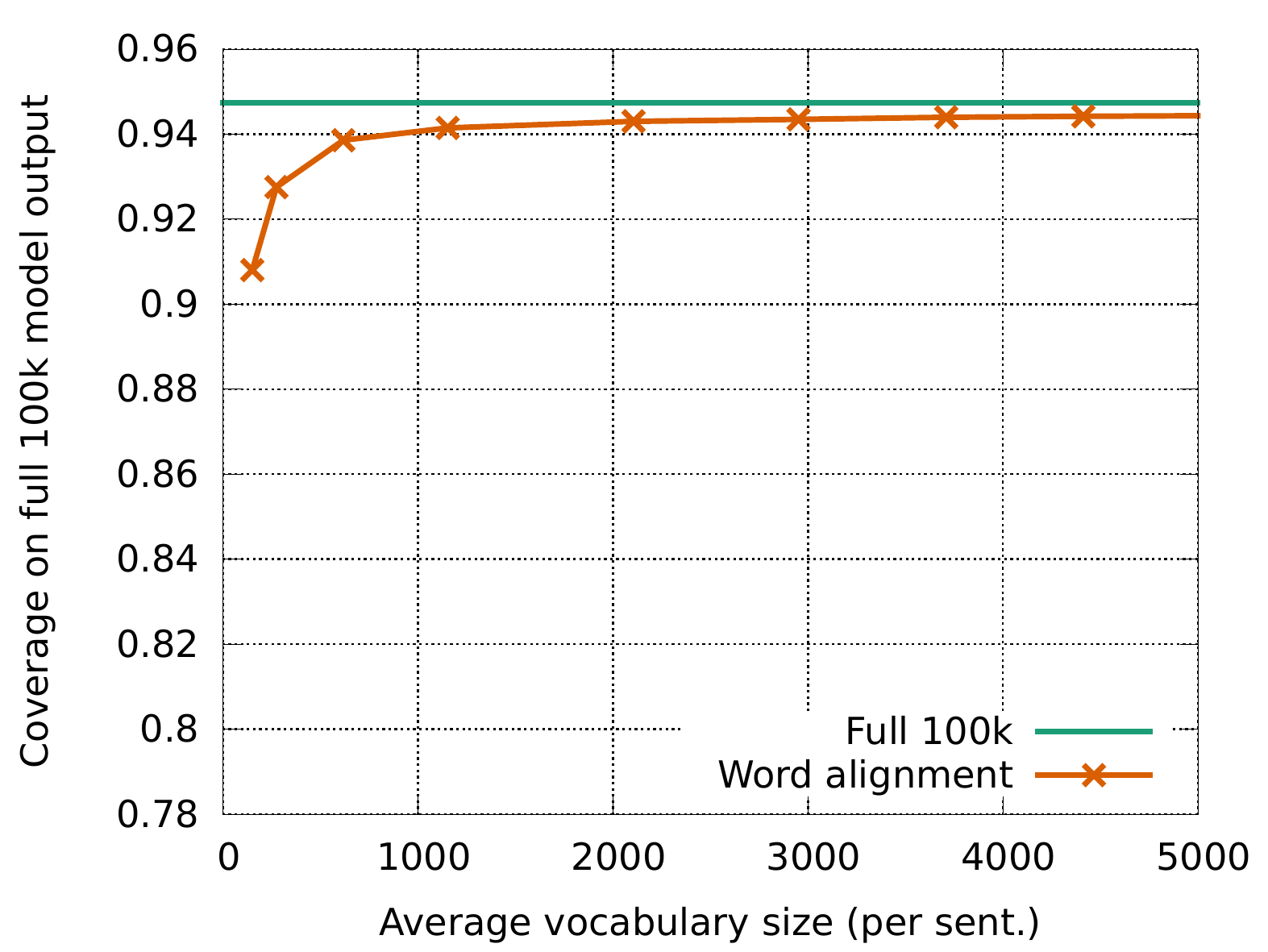}
\caption{\normalsize{
\emph{Left:} Coverage of the reference by the word alignment selection for different vocabulary sizes and by the full 100k model. \emph{Right:} Coverage of the full vocabulary model prediction by the word alignment selection method. Coverage does not count unknown words, therefore the full model has non-perfect coverage on itself.
Vocabulary selection never fully covers the reference (left) but it almost entirely covers the prediction of the full vocabulary model, even when very few candidates are selected.}}
\label{fig:coverage_vs_vocab}
\end{figure*}

Next we try to get a better sense of how precise selection
is with respect to the words used by a human translator or with 
respect to the translations generated by the full vocabulary model.
We use word alignments for this experiment.
Figure~\ref{fig:coverage_vs_vocab} shows coverage with respect to 
the reference (left) and with respect to the output of the full 
vocabulary system (right). 
We do not count unknown words (UNK) in all settings, even if they may
later be replaced by a source word (\textsection\ref{sec:expsetup}).
Not counting UNKs is the reason why the full vocabulary models do 
not achieve $100$\% coverage in either setting.
The two graphs show different trends: 
On the left, coverage with respect to the reference for the full 
vocabulary is $95.1$\%, while selection achieves $87.5$\%
with a vocabulary of 614 words (3rd point on graph).
However, when coverage is measured with respect to the full vocabulary
system output, the coverage of selection is very close to the full 
vocabulary model with respect to itself, i.e., when unknown words 
are not counted. 
In fact, the selection model covers over $99$\% of the non-UNK words 
in the full vocabulary output.
This result shows that selection can recover almost all of the 
words which are effectively selected by a full vocabulary model 
while discarding many words which are not chosen by the full model.

\begin{figure}[htb]
\centering
\includegraphics[width=\columnwidth]{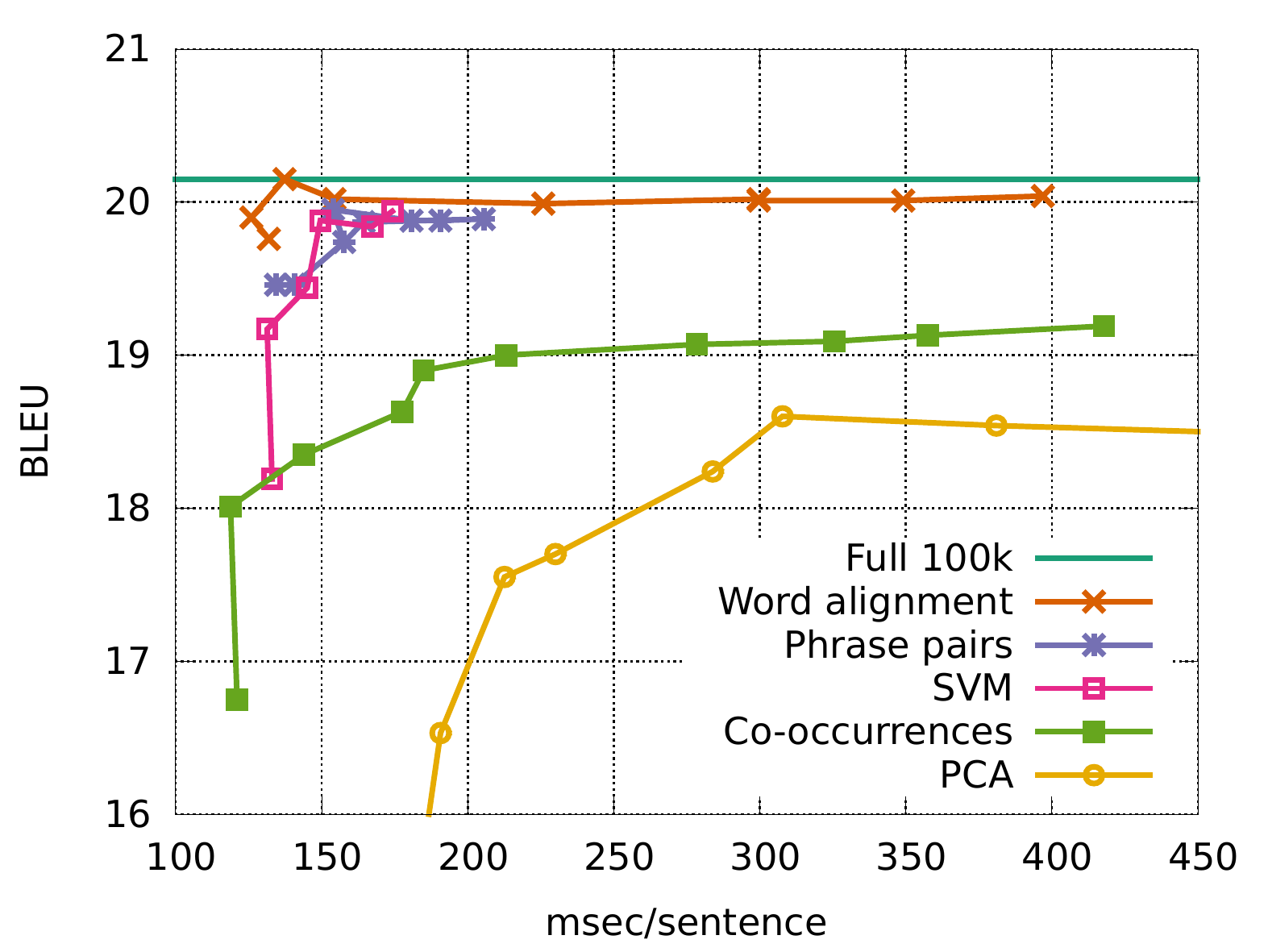}
\caption{\normalsize{BLEU accuracy versus decoding speed for a beam size of $5$ on CPU. Significant speed ups can be achieved with no decrease in BLEU accuracy, e.g., word alignment selection achieves $20.2$ BLEU at $137$ msec/sentence while the full vocabulary model requires $1,581$ msec/sentence at the same accuracy level, this is equivalent to an 11-fold speed up.}}
\label{fig:bleu_vs_decoding_time}
\end{figure}

\begin{table*}[htb]
\centering
\begin{tabular}{l|c cccccrcr }
{\bf EN-DE} & {\bf Param.} & {\bf 2010} & {\bf 2011} & {\bf 2012} & {\bf 2014} & {\bf 2015} & {\bf Voc.} & {\bf Cov.} & {\bf Time}  \\ \hline\hline
Full vocabulary & --      & 18.5 & 16.5 & 16.8 & 19.0 & 22.5 & 100,000 & 93.3\% & 1,524 \\ \hline
Co-occurrences  & top 300 & 17.2 & 15.6 & 15.8 & 18.1 & 20.6 & 1,036   & 81.1\% & 156  \\ \hline
PCA             & top 100 & 15.4 & 13.7 & 14.2 & 14.5 & 18.6 & 966     & 74.8\% & 143  \\ \hline
Word align      & top 100 & 18.5 & 16.4 & 16.7 & 19.0 & 22.2 & 1,093   & 88.5\% & 143  \\ \hline
Phrase pairs    & top 200 & 18.1 & 16.2 & 16.6 & 18.9 & 22.0 & 857     & 86.2\% & 153  \\ \hline
SVM             & --      & 18.3 & 16.2 & 16.6 & 18.8 & 21.9 & 1,284   & 86.6\% & -    \\\multicolumn{8}{c}{~}\\ 
{\bf EN-RO} & {\bf Param.} &&&&& {\bf 2016} & {\bf Voc.} & {\bf Cov.}  & {\bf Time}  \\ \hline\hline
Full vocabulary & --      &       &       &       &      & 27.9        & 50,000 & 96.0\% & 966 \\ \hline
Word align      & top 50  &       &       &       &      & 28.1        & 691    & 89.3\% & 186  \\
\end{tabular}
\caption{Final decoding accuracy results for WMT English-German and English-Romanian on various test 
sets (newstest 2010 -- 2016, except 2013 our validation set). We report the average vocabulary size 
per sentence, coverage of the reference and decoding time in milliseconds for newstest2015 and newstest2016. The parameter column indicates the maximum number of selected
candidates per source word or phrase.}
\label{tbl:test_bleu}
\end{table*}

What is the exact speed and accuracy trade-off when reducing the output 
vocabulary?
Figure~\ref{fig:bleu_vs_decoding_time} plots BLEU against decoding speed.
We pick a number of operating points from this graph
for our final test set experiments (Table~\ref{tbl:test_bleu}).
For our best methods (word alignments, phrase alignments and SVMs) 
we pick points such that vocabularies are kept small while maintaining
good accuracy compared to the full vocabulary setting. 
For co-occurrence counts and bilingual PCA we choose settings with 
comparable speed.

Our test results (Table~\ref{tbl:test_bleu}) confirm the validation 
experiments. 
On English-German translation we achieve more than a $10$-fold speed-up
over the full vocabulary setting. 
Accuracy for the word alignment-based selection matches the full
vocabulary setting on most test sets or decreases only slightly.
For example with word alignment selection, the largest drop is 
on newstest2015 which achieves $22.2$ BLEU compared to $22.5$ BLEU 
for the full setting on English-German; the best single-system neural 
setup at WMT15 achieved $22.4$ BLEU on this dataset~\cite{jean:wmt15}. 
On English-Romanian, we achieve a speed-up of over $5$ times with word 
alignments at $28.1$ BLEU versus $27.9$ BLEU for the full vocabulary
baseline.
This matches the state-of-the-art on this dataset~\cite{sennrich:wmt16}
from WTM16.
The smaller speed-up on English-Romanian is due to the smaller vocab of 
the baseline in this setting which is $50$k compared to $100$k for 
English-German.

\subsection{Selection for Better Training}

So far our evaluation focused on vocabulary selection for decoding, 
relying on a model trained with the full vocabulary. 
Next we address the question of whether the efficiency advantages 
observed for decoding translate to training as well. 
Selection at training may impact generalization performance 
either way:
it assimilates training and testing conditions which could positively 
impact accuracy. 
However, training with a reduced vocabulary could result 
in worse parameter estimates, especially for rare words which 
would receive much fewer updates because they would be selected 
less often.

We run training experiments on WMT English to German with
word alignment-based selection. 
In addition to the selected words, we include the target words
of the reference and train a batch with the union of the 
sentence-specific vocabularies of all samples~\cite{mi:manipulation}.

\begin{figure}[htb]
\centering
\includegraphics[width=\columnwidth]{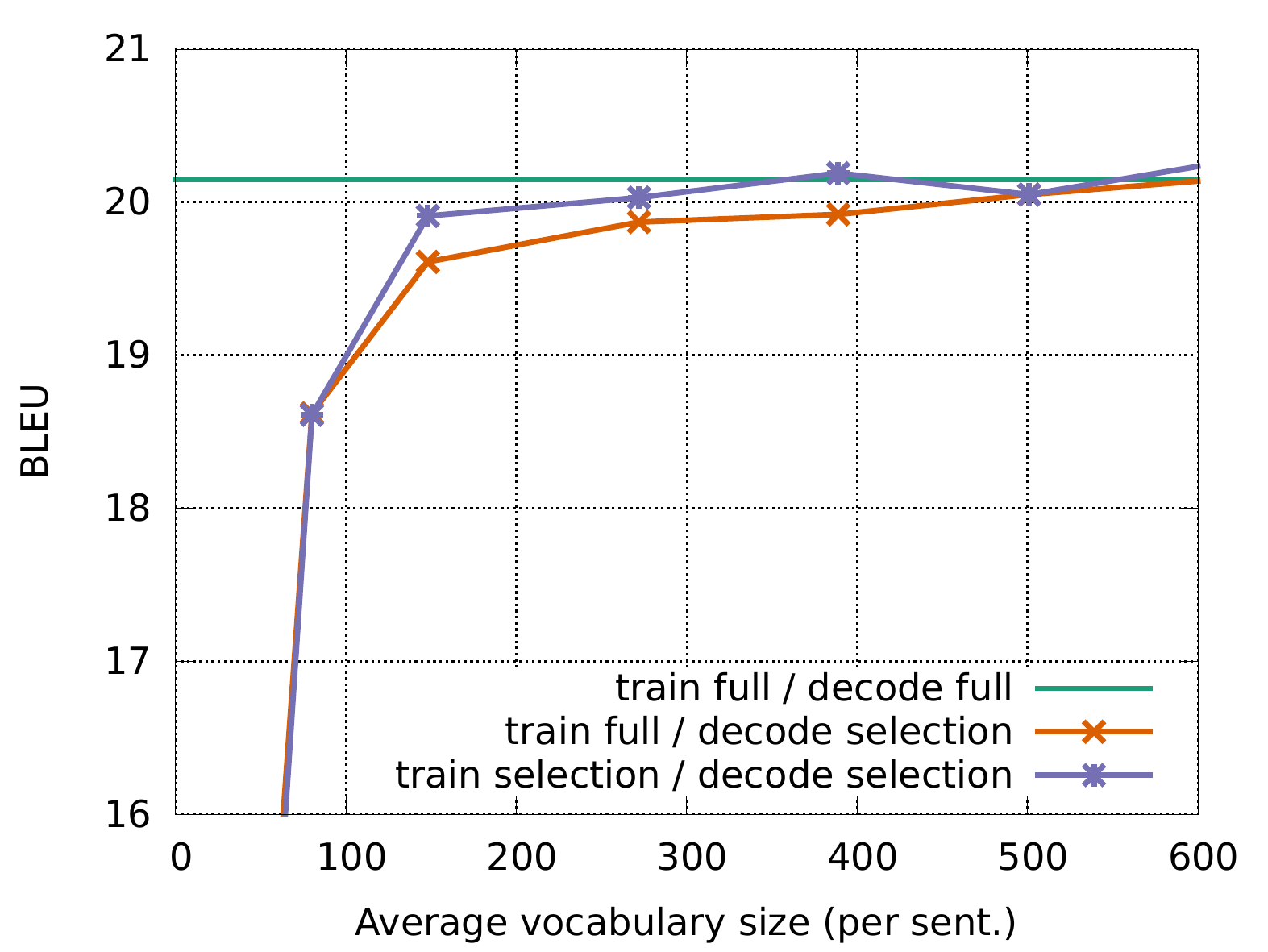}
\caption{\normalsize{
Accuracy on the validation set for different vocabulary sizes when using word alignment-based selection during training and testing, or the full vocabulary.}}
\label{fig:bleu_vs_vocab_train}
\end{figure}

Figure~\ref{fig:bleu_vs_vocab_train} compares validation accuracy
of models trained with selection or with the full vocabulary.
Selection in both training and decoding gives a small accuracy
improvement. However, this improvements disappears for 
vocabulary sizes of $500$ and larger; we found the same pattern 
on other test sets.
Similar to our decoding experiments, adding common words during 
training did not improve accuracy.



Table~\ref{tbl:train_time} shows the impact of selection on 
training speed.
Our bi-directional LSTM model (BLSTM) can process the same 
number of samples in 25\% less time on a GPU with a batch size 
of 32 sentences. 
We do not observe changes in the number of epochs required 
to obtain the best validation BLEU.

The speed-ups for training are significantly smaller than 
for decoding (Table~\ref{tbl:test_bleu}).
This is because training scores the vocabulary exactly once 
per target position, while beam search has to score multiple 
hypotheses at each generation step.

We suspect that training is now dominated by the bi-directional LSTM
encoder. 
To confirm this, we replaced the encoder with a simple average pooling
model which encodes source words as the mean of word and 
position embeddings over a local context \cite{ranzato:2015:iclr}.
Table~\ref{tbl:train_time} shows that in this setting the efficiency
gains of vocabulary selection are more substantial 
($40\%$ less time per epoch). 
This model is not as accurate and achieves only $18.5$ BLEU on 
newstest2015 compared to $22.5$ for the bi-directional LSTM encoder. 
However, it shows that improving the efficiency of the encoder is 
a promising future work direction.

\begin{table}[htb]
\centering
\begin{tabular}{l | c c c}
{\bf Vocab. per batch}   & {\bf 100k} & {\bf 6k} &        \\\hline\hline
Avg. pooling encoder   &  5h 55 & 3h 34 & (-40\%)\\
BLSTM encoder            & 9h 34 & 7h 13 & (-25\%)\\
\end{tabular}
\caption{%
Training times per epoch over $3.6$m sentences in hours and minutes 
on German-English for the full ($100$k) and reduced vocabulary 
settings ($6$k).
Measurements include forward/backward/update on a GPU for a batch 
of size $32$. The $6$k candidate words per batch correspond to an 
average of $390$ words per sentence.
}
\label{tbl:train_time}
\end{table}

\section{Conclusions}
\label{sec:ccl}

This paper presents a comprehensive analysis of vocabulary 
selection techniques for neural machine translation. 
Vocabulary selection constrains the output words to be scored 
to a small subset relevant to the current source sentence.
The idea is to avoid scoring a high number of unlikely 
candidates with the full model which can be ruled out 
by simpler means.

We extend previous work by considering a wide range of 
simple and complex selection techniques including 
bilingual word co-occurrence counts, bilingual embeddings 
built with Hellinger PCA, word alignments, phrase pairs,
and discriminative SVM classifiers.
We explore the trade-off between speed and accuracy for 
different vocabulary sizes and validate results on two 
language pairs and several test sets.

Our experiments show that decoding speed-up can be
reduced by up to $90$\% without compromising accuracy. 
Word alignments, bilingual phrases and SVMs can achieve 
high accuracy, even when considering fewer than $1,000$ 
word types per sentence.

At training time, we achieve a speed-up of up to $1.33$ with
a bi-directional LSTM encoder and $1.66$ with a faster
alternative.
Efficiency increases are less pronounced during training 
because of two combined factors.
First, vocabulary scoring at the final layer of the model 
is a smaller part of the computation compared to beam search.
Second, state-of-the-art bi-directional LSTM 
encoders~\cite{bahdanau:iclr15} are relatively costly 
compared to scoring the vocabulary on GPU hardware.
Efficiency gains from vocabulary selection highlight 
the importance of progress towards efficient, accurate 
encoder and decoder architectures.

\bibliography{vocabulary-selection.bib}
\bibliographystyle{eacl2017}

\end{document}